\documentclass[letterpaper, 10 pt, conference]{ieeeconf}  

\IEEEoverridecommandlockouts                              

\usepackage{graphicx}
\usepackage{capt-of}
\usepackage[caption=false]{subfig}
\usepackage{amsfonts,amssymb,amsbsy,amsmath}
\usepackage{booktabs,multirow,multicol,makecell}
\makeatletter
\let\NAT@parse\undefined
\makeatother
\usepackage{hyperref}

\usepackage{lipsum}
\makeatletter
\newcommand{\showfontsize}{\f@size pt}
\makeatother

\newcommand{\trans}{T}

\newcommand{\bsym}[1]{\boldsymbol{#1}}

\newcommand{\dbsym}[1]{\dot{\boldsymbol{#1}}}

\newcommand{\mbf}[1]{\mathbf{#1}}

\newcommand{\dmbf}[1]{\dot{\mathbf{#1}}}

\newcommand{\pth}[1]{\left(#1\right)}

\newcommand{\norm}[1]{\left\|#1\right\|}
\newcommand{\bmat}[1]{\begin{bmatrix}#1\end{bmatrix}}

\title{\LARGE \bf
PccDiffuser: Multi-solution Motion Planning for Continuum Robots
}

\author{Ke Qiu$^*$, Sifan Chen$^*$, Si Wang, Rong Xiong, Yue Wang and Haojian Lu$^\dagger$
\thanks{This work was supported by ...}
\thanks{The authors are with Department of Control Science and Engineering, Zhejiang University, Hangzhou, China.}
\thanks{$^*$Equal contribution. }
\thanks{$^\dagger$Corresponding author: Haojian Lu (e-mail: luhaojian@zju.edu.cn).}
}

\begin{document}

\IEEEaftertitletext{\begin{minipage}{\textwidth}
  \centering
  \includegraphics[width=7.0in]{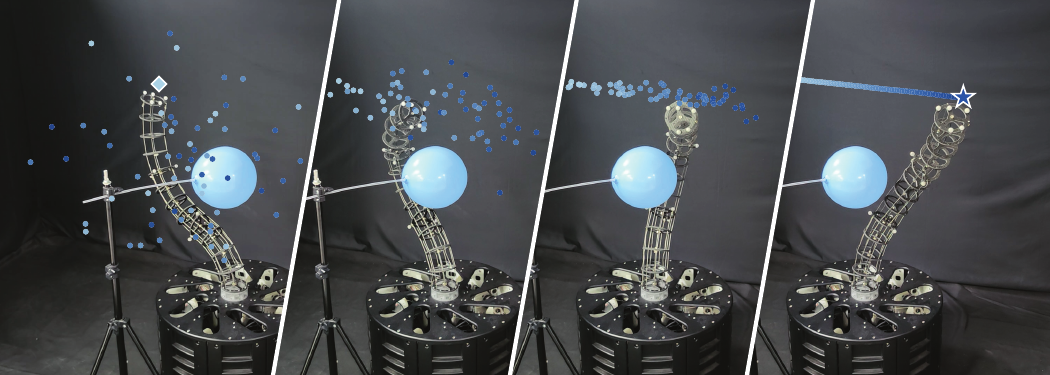}
  \captionof{figure}{Obstacle-aware planning and hardware execution. From left to right, the candidate tip paths at different denoising steps are overlaid on the robot workspace. The samples progressively concentrate into feasible motions from the initial configuration (diamond) to the target tip position (star), and the executed final trajectory passes around the ball obstacle without contact. (See also the supplementary video.)}
  \label{fig:teaser}
  \vspace{\dbltextfloatsep}
\end{minipage}}

\maketitle
\thispagestyle{empty}
\pagestyle{empty}

\begin{abstract}
We present the PccDiffuser, a conditional diffusion framework for continuum robots that learns a multimodal distribution over complete configuration-space paths and samples multiple candidate solutions in parallel, which are subsequently converted into an executable trajectory by time allocation considering actuator constraints. Under the piecewise constant-curvature model, we use exponential co-ordinates to describe the robot kinematics, and use graph neural network to encode a variable number of environment obstacles. Analytical differential kinematics is incorporated in the denoising process to improve terminal accuracy and whole-body clearance. On a mixed test set comprising workspace with zero to four obstacles, PccDiffuser achieved a success rate of 91\%. Compared with existing sampling- and optimisation-based benchmarks, it delivered both a higher success rate and greater computational efficiency, with the latter advantage becoming more substantial when sampling more candidate solutions. Experiments on a three-section tendon-driven continuum robot further demonstrate consecutive planning, multi-solution planning, and whole-body obstacle avoidance.
\end{abstract}

\section{INTRODUCTION}

Continuum robots are continuously deformable structures that are complient and adaptive in cluttered environments~\cite{rus2015design}. These properties grant them promising applications such as minimally invasive surgery~\cite{burgner2015continuum}, industrial inspection and maintenance~\cite{teetaert2026continuum,yang2024novel}, and human-robot interaction~\cite{abah2021multi}.

For continuum robots, task-specific requirements are generally imposed on path geometry, whereas a selected path is typically expected to be executed in the shortest feasible time. We therefore separate motion planning into two stages: first generating feasible paths and then allocating time under actuator constraints to construct executable trajectories. During path planning, kinematic redundancy permits multiple terminal configurations and substantially different whole-body paths to reach the same target~\cite{qiu2023efficient}. This multiplicity provides useful alternatives for obstacle avoidance or inference-time objectives at runtime, but also makes the planning problem underconstrained and inherently multimodal.

Existing sampling- and optimisation-based planners search either the workspace or the configuration space~\cite{meng2022rrt,gao2024study}. They generally solve each case through an iterative search and return only one solution per run. Consequently, they do not naturally capture the multimodality of the problem. Configuration-space planners explore the search space with limited task guidance. Moreover, due to the non-linear kinematics of continuum robots, a path in configuration space may produce an undesirable path in the workspace~\cite{meng2022rrt}. Workspace planners describe tip motion more naturally, but repeatedly solving inverse kinematics can produce discontinuities or converge to local minima~\cite{carvalho2025motion}. A planner that efficiently generates diverse solutions while maintaining terminal accuracy, high-quality workspace path, and whole-body obstacle avoidance is therefore desirable.

Diffusion models offer a promising way to address this challenge~\cite{ho2020denoising}. By learning a conditional distribution, they can represent multimodal actions and generate several candidates in parallel from different initial noise samples~\cite{carvalho2025motion,janner2022planning,chi2025diffusion}. Their successful use in rigid manipulators suggests that the same principle could benefit continuum robots. Can we bring the ability of diffusion models to the planning of continuum robots?

In this paper, we present PccDiffuser, a conditional diffusion framework for multi-solution motion planning of continuum robots in piecewise constant curvature (PCC) kinematics. As shown in Fig.~\ref{fig:teaser}, it generates configuration-space paths from Gaussian noise, conditioned on initial configuration, target tip position, and surrounding obstacles. A graph neural network (GNN) accommodates variable number of obstacles, while analytical differential kinematics guides the reverse process to improve terminal accuracy and whole-body clearance. Selected paths are subsequently converted into executable trajectories via time allocation considering actuator constraints. Batched denoising allows multiple candidate solutions to be generated simultaneously, making the exploration of alternative motions computationally practical. The contributions are summarised as follows.
\begin{itemize}
  \item We propose a diffusion motion planner for continuum robots that can generate multiple configuration-space solutions in parallel, conditioned on initial configuration, target position, and obstacles.
  \item We validate the planner in simulation and on a three-section tendon-driven continuum robot, demonstrating its multimodalilty, improved success rate and efficiency.
\end{itemize}

\section{RELATED WORK}

In this section, we review modelling and planning methods for continuum robots, then discuss diffusion models and their applications in robot motion planning.

\subsection{Modelling and Planning}

The piecewise constant curvature model describes the robot as concatenated circular arcs and provides explicit forward kinematics evaluation~\cite{webster2010design}. Higher-fidelity Cosserat rod models represent distributed strain and more general deformation, but require greater computational cost~\cite{mathew2025reduced}. Consequently, motion planners commonly adopt the PCC model to support their search. Even with this reduced representation, however, a desired tip position generally admits multiple terminal configurations and distinct backbone shapes~\cite{qiu2023efficient}. Planning must therefore consider the entire robot body and the multiplicity of inverse kinematics solutions.

Rapidly-exploring random trees (RRT) and its variants are widely used in continuum robot planning. Classified by the search space, workspace methods find a tip path and recover robots configuration at waypoints through inverse kinematics~\cite{meng2022rrt}. Obstacle avoidance objectives can be incorporated into the optimisation for whole-body safety~\cite{luo2024efficient}. Nevertheless, the planning process depends on frequent successful iterations, which may converge to local minima or become discontinuous between adjacent nodes. Configuration-space methods instead grow a tree directly among whole-body PCC configurations~\cite{gao2024study}, where forward kinematics enables whole-body collision checking as the trees grow. Cross-entropy sampling has been used to improve the sampling stage of RRT*~\cite{chen2023cross}. These methods, however, receive limited direct guidance from the workspace task, and the non-linear kinematic mapping means that a short configuration-space path may produce an undesirable workspace path~\cite{meng2022rrt}. Moreover, both planners normally return one solution per run. Representing multiple feasible motion modes requires repeated searches. This limitation motivates learning a generative distribution over valid and collision-free paths.

\subsection{Diffusion Models}

The denoising diffusion probabilistic model (DDPM) learns to reverse a gradual noising process. A network begins generation from noisy samples and proceeds through iterative denoising~\cite{ho2020denoising}. The denoising diffusion implicit model (DDIM) use the same trained denoiser with a shorter, optionally deterministic sampling schedule~\cite{song2020denoising}. Different initial noise samples can therefore explore distinct modes, while deterministic sampling makes each solution reproducible.

These properties have enabled diffusion models to generate actions and complete paths. Trajectory-level diffusion has been applied to rigid manipulators, with goal conditions and task objectives incorporated during denoising~\cite{janner2022planning}. Multimodal visuomotor action sequences are generated through conditional denoising~\cite{chi2025diffusion}. Path priors are learned for planar robots and serial manipulators and applies collision and goal guidance during sampling~\cite{carvalho2025motion}. Diffusion action generation has also been incorporated into vision-language-action policies for cross-embodiment, short-horizon end-effector prediction~\cite{chen2025toward}. Recent work has extended environment-aware diffusion planning to the backbone shapes of hybrid rigid-soft manipulators~\cite{kamtikar2026thread}. Nevertheless, multimodal path generation, conditioned on initial configuration, tip target and a variable number of obstacles, remains underexplored. This paper addresses this problem by learning complete configuration-space paths and incorporating analytical continuum robot kinematics during sampling.

\section{PROBLEM FORMULATION}

In this section, we introduce a non-singular representation for the kinematics of multi-section PCC robots, then formulate the trajectory planning problem and separates it into geometric path generation and time allocation.

\subsection{Non-Singular Kinematics Representation}

We consider a three-section PCC robot. For section $i \in \{1,2,3\}$, let $L_i>0$ denote the fixed arc length, $\kappa_i\geq0$ the constant curvature, and $\phi_i\in[-\pi,\pi)$ the angle of bending plane about the local $z$-axis. Although $(\kappa_i,\phi_i)$ is geometrically intuitive, this parameterisation suffers from discontinuity and singularity at the straight configuration~\cite{della2020improved}. Therefore, we adopt the exponential co-ordinates of the PCC model~\cite{webster2010design}. Using an angular-first twist convention, the integrated spatial twist of the $i$-th section is given by
\begin{equation}
  \bsym{\xi}_i = L_i \bmat{-\kappa_i\sin\phi_i & \kappa_i\cos\phi_i & 0 & 0 & 0 & 1}^{\trans}.
\end{equation}
We use $\xi_{i1}$ and $\xi_{i2}$ to denote the first and second entries of the exponential co-ordinate. This parameterisation is equivalent and non-singular since
\begin{equation}
  \kappa_i=\frac{1}{L_i}\sqrt{\xi_{i1}^{2}+\xi_{i2}^{2}}, \quad \phi_i=\mathrm{atan2}(-\xi_{i1},\xi_{i2}).
\end{equation}
To prevent excessive bending, we constrain the curvature of each section such that $\kappa_i \leq \kappa_{\max}$. Finally, the robot configuration is formed by concatenating the variable entries of the twist for all sections,
\begin{equation}
  \bsym{\zeta}
  = \bmat{\xi_{11} & \xi_{12} & \xi_{21} & \xi_{22} & \xi_{31} & \xi_{32}}^{\trans}
  \in\mathbb{R}^{6}.
\end{equation}
The transformation $\mbf{T} \in \mathrm{SE}(3)$ from the base to the robot tip is then obtained using the product of exponentials~\cite{murray1994mathematical},
\begin{equation}
  \mbf{T}(\bsym{\zeta}) = \exp\pth{\bsym{\xi}^\wedge_1} \exp\pth{\bsym{\xi}^\wedge_2} \exp\pth{\bsym{\xi}^\wedge_3}.
\end{equation}
The Jacobian follows the standard definition~\cite{lynch2017modern}. Let $\dbsym{\zeta}$ and $\dmbf{T}$ denote their time derivative, then
\begin{equation}
  \mbf{J} \pth{\bsym{\zeta}} \dbsym{\zeta} = \mbf{T}^{-1} \dmbf{T}.
\end{equation}
An expression for $\mbf{J}(\bsym{\zeta})$ is derived in~\cite{chirikjian2011stochastic}, and we extract its translational component $\mathrm{d} \mathcal{P}(\bsym{\zeta}) / \mathrm{d} \bsym{\zeta}$, where $\mathcal{P}(\bsym{\zeta})$ denotes the tip position. This supports denoising guidance using analytical differential kinematics in sampling stages.


\subsection{Obstacle-Aware Trajectory Planning}

We model the environment obstacles as a collection of balls $\mathcal{O} = \bigcup_{i \ge 0} B(\mbf{c}_i,r_i)$, where $B(\mbf{c}_i,r_i) = \{\mbf{y} \in \mathbb{R}^{3}: \norm{\mbf{y} - \mbf{c}_i} \leq r_i\}$, $\mbf{c}_i \in \mathbb{R}^{3}$ is the centre, and $r_i>0$ is the radius of the $i$-th obstacle. Let $\mathcal{B}(\bsym{\zeta})\subset\mathbb{R}^{3}$ denote the volume occupied by the entire robot body. Given an initial configuration $\bsym{\zeta}^{*}_{0}$ and a target tip position $\mathcal{P}^{*}$, the planning problem is to find $\bsym{\zeta}(v), v \in [0,V]$, such that
\begin{equation}
  \begin{aligned}
    &\bsym{\zeta}(0)=\bsym{\zeta}^{*}_{0},\\
    &\mathcal{P}(\bsym{\zeta}(V)) = \mathcal{P}^{*},\\
    &\mathcal{B}(\bsym{\zeta}(v)) \cap \mathcal{O}=\emptyset\quad(\forall v).
  \end{aligned}
\end{equation}
We use $v \in [0,V]$, where $V$ is the total duration, for physical time, reserving $t$ for the diffusion step.

We separate the problem into geometric path planning and subsequent time scheduling. For static environments, test-time preferences are generally imposed on the path, while a shorter execution time is preferred for a selected path. Geometric requirements such as the terminal condition and whole-body collision avoidance are invariant under any time parameterisation. Specifically, we discretise a path at $H$ configurations $\bsym{\zeta}_0, \bsym{\zeta}_1, \dots, \bsym{\zeta}_{H-1}$ and associate them with a time schedule $v_h, h = 0, 1, \dots, H - 1$. The path planner provides multiple candidates satisfying
\begin{equation}
  \begin{aligned}
    &\bsym{\zeta}_0=\bsym{\zeta}^{*}_{0},\\
    &\mathcal{P}(\bsym{\zeta}_{H-1}) = \mathcal{P}^{*},\\
    &\mathcal{B}(\bsym{\zeta}_h) \cap \mathcal{O} = \emptyset\quad(\forall h),
  \end{aligned}
\end{equation}
We assume that adjacent configurations are sufficiently close that, if both are collision-free, the configurations interpolated between them are also collision-free. Each candidate is then assigned a minimum-duration schedule $0=v_0<\dots<v_{H-1}$ subject to the actuation velocity constraints.

Consequently, the path planner generates multiple feasible candidates, and subsequent time allocation converts a selected path into an executable trajectory. Together, these two components constitute the complete trajectory planner.

\section{PLANNING WITH CONDITIONAL DIFFUSION}

This section presents PccDiffuser, which learns a conditional distribution over complete configuration-space paths, draws multiple solutions from it in parallel, and allocate time to convert paths into trajectories.

\subsection{Conditional Diffusion Model}
Let $\mbf{x}_{0}$ denote a feasible configuration-space path and $\mbf{c}$ the given planning condition including the initial configuration, target tip position, and obstacles,
\begin{equation}
  \begin{aligned}
    &\mbf{x}_{0} = \bmat{\bsym{\zeta}_0&\dots&\bsym{\zeta}_{H-1}}^{\trans} \in \mathbb{R}^{H \times 6},\\
    &\mbf{c} = \{\bsym{\zeta}^{*}_{0},~\mathcal{P}^{*},~\mathcal{O}\}.
  \end{aligned}
\end{equation}
Let $\mbf{x}_t$ be its noisy state at diffusion step $t = 1, \dots, T$. The conditional diffusion model learns $p_{\theta}(\mbf{x}_{t-1} | \mbf{x}_{t}, \mbf{c})$ by pairing a fixed forward process that adds Gaussian noise to the data with a conditional reverse process that removes this noise from $\mbf{x}_T \sim \mathcal{N}(\mbf{0},\mbf{I})$~\cite{ho2020denoising},
\begin{equation}
  p_{\theta}(\mbf{x}_{0:T} | \mbf{c}) = p(\mbf{x}_T)\prod_{t=1}^{T} p_{\theta}(\mbf{x}_{t-1} | \mbf{x}_t,\mbf{c}).
\end{equation}
All path and condition variables are normalised before being processed by the diffusion model.


\textbf{Forward Process.} In this stage, the path is progressively corrupted according to
\begin{equation}
  q\pth{\mbf{x}_{t} | \mbf{x}_{t-1}} = \mathcal{N}\pth{\sqrt{\alpha_t}\mbf{x}_{t-1},~(1-\alpha_t)\mbf{I}},
\end{equation}
where $\alpha_t \in (0,1)$. Defining $\bar{\alpha}_t=\prod_{j=1}^{t}\alpha_j$ permits direct sampling at an arbitrary diffusion step, and we use the cosine schedule~\cite{nichol2021improved} to determine $\bar{\alpha}_t$ and the variance,
\begin{equation}
  \mbf{x}_{t} = \sqrt{\bar{\alpha}_t}\mbf{x}_{0} + \sqrt{1-\bar{\alpha}_t}\bsym{\epsilon},\quad\bsym{\epsilon}\sim\mathcal{N}\pth{\mbf{0},\mbf{I}}.
\end{equation}

\textbf{Reverse Process.} Rather than predicting the reverse mean and covariance independently, the conditional denoiser predicts the noise contained in the current path, i.e., $\bsym{\epsilon}_{\theta}\pth{\mbf{x}_{t},t,\mbf{c}}$. We employ deterministic DDIM sampling ($\eta = 0$)~\cite{song2020denoising}. Let $s < t$ denote the next selected diffusion step in the respaced sampling schedule, then the DDIM update gives
\begin{equation}
  \begin{aligned}
    \mbf{x}_{s} = \sqrt{\bar{\alpha}_{s}} \cdot \frac{1}{\sqrt{\bar{\alpha}_{t}}} \pth{\mbf{x}_{t}-\sqrt{1-\bar{\alpha}_{t}} \bsym{\epsilon}_{\theta}(\mbf{x}_{t},t,\mbf{c})}&\\
    {} + \sqrt{1-\bar{\alpha}_{s}} \bsym{\epsilon}_{\theta}(\mbf{x}_{t},t,\mbf{c})&.
  \end{aligned}
\end{equation}
Starting from $\mbf{x}_{T}\sim\mathcal{N}(\mbf{0},\mbf{I})$, this update is applied over 50 steps selected from the $T=1000$ training steps to produce a path conditioned on the planning inputs.

\subsection{Network Architecture}

\begin{figure}[t]
  \centering
  \includegraphics[width=3.4in]{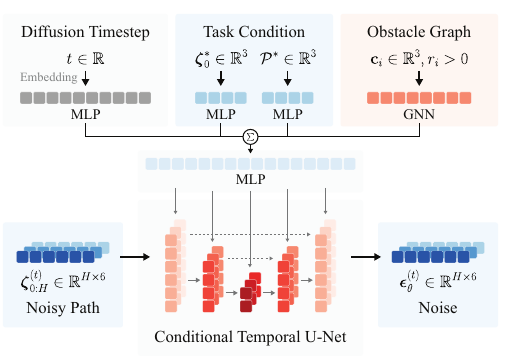}
  \caption{The diffusion step, initial configuration, target position, and obstacle graph are embedded into a shared conditioning vector, while a temporal U-Net predicts the noise given the configuration-space path.}
  \label{fig:network}
\end{figure}

As illustrated in Fig.~\ref{fig:network}, the denoiser is a one-dimensional temporal U-Net that receives the noisy path together with a binary channel identifying the inpainted initial configuration. The diffusion step, initial configuration, and target position are embedded by multilayer perceptrons, while a graph neural network (GNN) encodes a variable number of obstacles as a complete directed graph with $[\mbf{c}_i^{\trans},r_i]^{\trans}$ as node features and relative centre information in its messages. Graph pooling converts variable number of obstacles into a fixed-dimensional embedding, which is fused with the other conditions into a shared context vector and modulates the residual blocks of the U-Net through feature-wise affine transformations. The network consequently preserves temporal path structure while conditioning every resolution on the complete planning task. Rather than predicting the mean of the reverse transition, it predicts the noise $\bsym{\epsilon}_{\theta}(\mbf{x}_t,t,\mbf{c})$ for all 6 configuration components at each frame.

\subsection{Training}

\textbf{Training Data.} We generate configuration-space paths for a three-section PCC robot with length $L_i = 1$ and bending limit $\kappa_{\max} = \pi/2$, with $H = 64$ waypoints as training data. Tip targets are obtained from randomly sampled configurations, after which multiple terminal inverse kinematics solutions and initial configurations are paired. A straight workspace tip path is mapped into configuration space using damped differential kinematics with a null-space continuity term. In obstacle scenes, colliding paths are repaired by applying repulsion of an artificial potential field (APF) to backbone points, then using Douglas--Peucker simplification~\cite{visvalingam1990douglas}, and finally fitting a spline-interpolated workspace path. Those paths satisfying the bending, continuity, and clearance constraints are retained. We generate an obstacle-free dataset of around 353k paths and a mixed dataset of around 243k paths containing zero to four ball obstacles. Targets are divided into the training and test sets.

\textbf{Training Loss.} At each update, a diffusion step $t$ is sampled uniformly and Gaussian noise $\bsym{\epsilon}$ is added to a clean path. Let $\mbf{M} \in \{0, 1\}^{H\times6}$ be zero at the first configuration and one elsewhere. Since the initial configuration is restored by inpainting, it is excluded from the noise-prediction loss. Therefore, we use a masked conditional variant of the simplified noise-prediction objective~\cite{ho2020denoising}, i.e.,
\begin{equation}
  \mathcal{L} = \frac{1}{{6(H-1)}}\mathbb{E}_{\mbf{x}_0,t,\bsym{\epsilon}} \left[ \norm{\mbf{M} \odot \pth{\bsym{\epsilon}-\bsym{\epsilon}_{\theta}(\mbf{x}_t,t,\mbf{c})}}^2 \right],
\end{equation}
where $\odot$ denotes element-wise multiplication. All variables are normalised before training. We optimise the model with AdamW for 100k updates and use an exponential moving average of its parameters for inference.

\subsection{Sampling Guided by Differential Kinematics}
\label{subsec:sampling_guided_by_differential_kinematics}
\begin{figure}[t]
  \centering\vspace{-10pt}
  \subfloat[]{\includegraphics[width=1.7in]{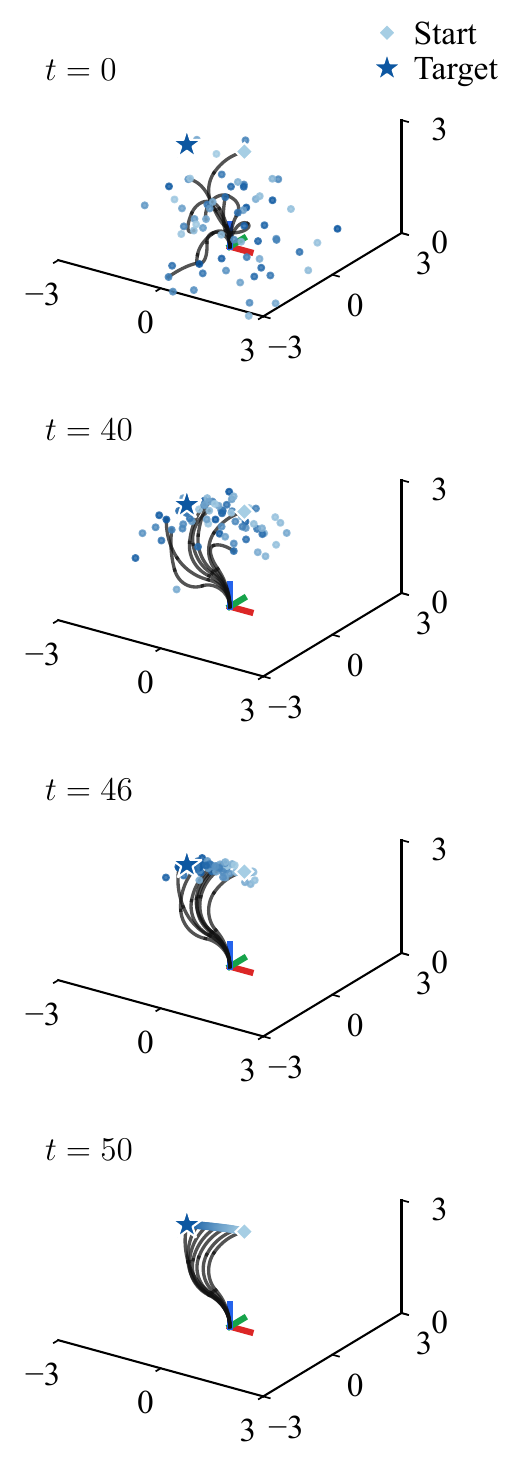}\label{fig:denoising_a}}
  \subfloat[]{\includegraphics[width=1.7in]{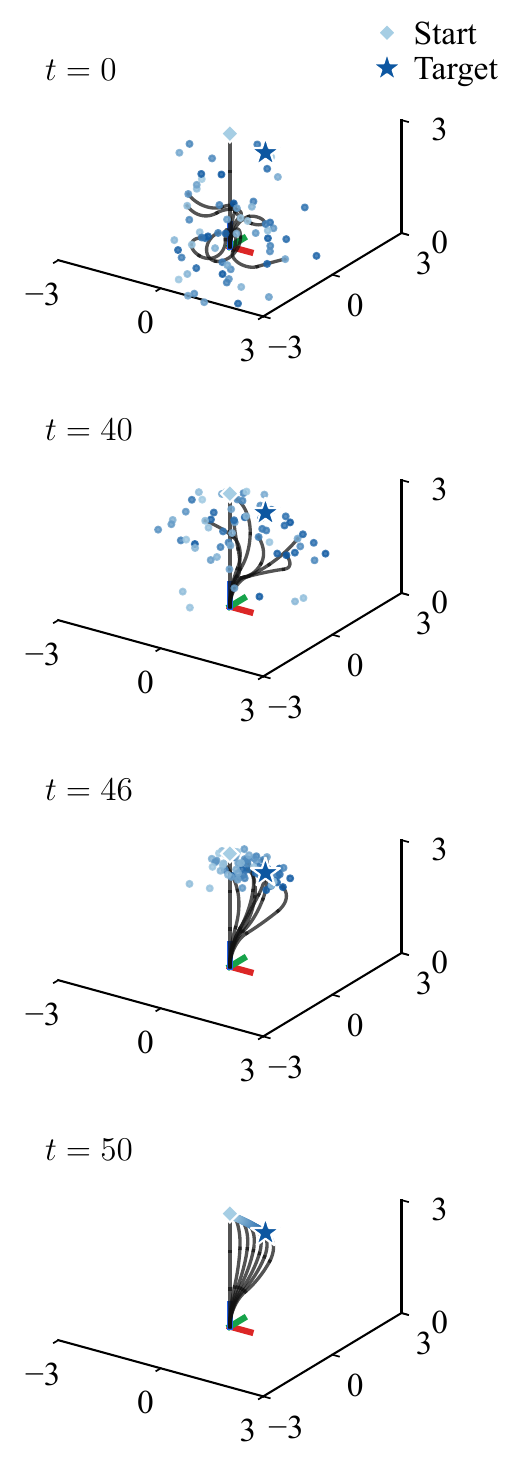}\label{fig:denoising_b}}
  \caption{Snapshots of path evolution during the 50-step DDIM reverse process ($T=1000$), starting from (a) a random initial configuration and (b) the straight configuration.}
  \label{fig:denoising}
\end{figure}

Fig.~\ref{fig:denoising} visualises how initially unstructured samples converge to feasible paths under the reverse process. During the process, analytical differential kinematics can refine task satisfaction. For a predicted path $\hat{\mbf{x}}_0$, the terminal objective and its gradient are
\begin{equation}
  \begin{aligned}
    &E_{1} = \frac{1}{2}\norm{\mathcal{P}(\bsym{\zeta}_{H-1})-\mathcal{P}^{*}}^2,\\
    &\nabla_{\bsym{\zeta}_{H-1}} E_{1} = \mbf{J}(\bsym{\zeta}_{H-1})^{\trans} \pth{\mathcal{P}(\bsym{\zeta}_{H-1})-\mathcal{P}^{*}},
  \end{aligned}
\end{equation}
where we abuse the notation $\mbf{J}(\bsym{\zeta}) = \mathrm{d}\mathcal{P}/\mathrm{d}\bsym{\zeta}$ to represent the translational Jacobian. We sampled points along the backbone and for the deepest penetrating point we compute the displacement $\Delta\mathcal{P}_k$ from prescribed safe boundary to that point. The corresponding configuration-space repulsion is
\begin{equation}
  \nabla_{\bsym{\zeta}_h} E_{2} = \mbf{J}_{k}(\bsym{\zeta}_h)^{\trans} \Delta\mathcal{P}_k,
\end{equation}
where $\mbf{J}_{k}(\bsym{\zeta}_h)$ is the Jacobian at the $k$-th selected backbone point on the $h$-th configuration. The update of $\bsym{\zeta}_0$ is always set to zero. We consider the following two methods for incorporating analytical guidance in the reverse process.

\textbf{Post Correction.} At step $t$, the predicted noise is first converted into the final path estimate $\mbf{x}^{(t)}_0$, then the terminal configuration is moved along $-\nabla_{\bsym{\zeta}_{H-1}}E_1$, while each penetrating configuration is moved along the repulsion direction $-\nabla_{\bsym{\zeta}_h}E_2$. Hence, for every row $\bsym{\zeta}^{(t)}_h$ of $\mbf{x}^{(t)}_0$, we apply
\begin{equation}
  \bsym{\zeta}^{(t)}_{h} \leftarrow
  \begin{cases}
    \bsym{\zeta}^{(t)}_{h} - w_{1} \nabla_{\bsym{\zeta}_{h}} E_{1} - w_{2} \nabla_{\bsym{\zeta}_{h}} E_{2},\quad&\text{if}~h = H-1,\\[2pt]
    \bsym{\zeta}^{(t)}_{h} - w_{2} \nabla_{\bsym{\zeta}_{h}} E_{2},\quad&\text{otherwise}.
  \end{cases}
\end{equation}
The corrected $\mbf{x}^{(t)}_0$ is then substituted into the DDIM update to obtain the next noisy state $\mbf{x}_s$, where $s < t$.

\textbf{Guided Prediction.} Guided prediction instead modifies the predicted noise. Let $E=w_1E_1+w_2E_2$ be evaluated on the predicted final path
\begin{equation}
  \label{eq:x^t_0}
  \mbf{x}^{(t)}_0 = \frac{\mbf{x}_t - \sqrt{1 - \bar{\alpha}_t} \bsym{\epsilon}_{\theta}(\mbf{x}_t, t, \mbf{c})}{\sqrt{\bar{\alpha}_t}}.
\end{equation}
Since this estimate depends on $\mbf{x}_t$ through the denoiser, we use automatic differentiation propagates the analytical geometric gradient through~(\ref{eq:x^t_0}) to obtain $\nabla_{\mbf{x}_t}E$. The guided noise is given by
\begin{equation}
  \bsym{\epsilon}_\theta \leftarrow \bsym{\epsilon}_{\theta} + \sqrt{1-\bar{\alpha}_t} \nabla_{\mbf{x}_t} E,
\end{equation}
which is subsequently used in the DDIM update.

\subsection{Trajectory Generation}

After sampling a configuration-space path, we can obtain its corresponding actuator-space path using a linear mapping, and assign minimal time subject to actuator velocity and acceleration limits~\cite{qiu2025actuator}. The scheduling preserves the planned geometry, and the resulting trajectory can be spline-interpolated when executed on the hardware platform.

\section{EXPERIMENTS}

We evaluate the proposed PccDiffuser in both simulation and real world. The experiments examine the ability to generate multiple feasible solutions, quantify the effects of sampling guidance, compare its performance with conventional planners, and demonstrate real-world execution.

\subsection{Multi-Solution Planning}

\begin{figure}[t]
  \centering
  \subfloat[]{\includegraphics[width=3.4in]{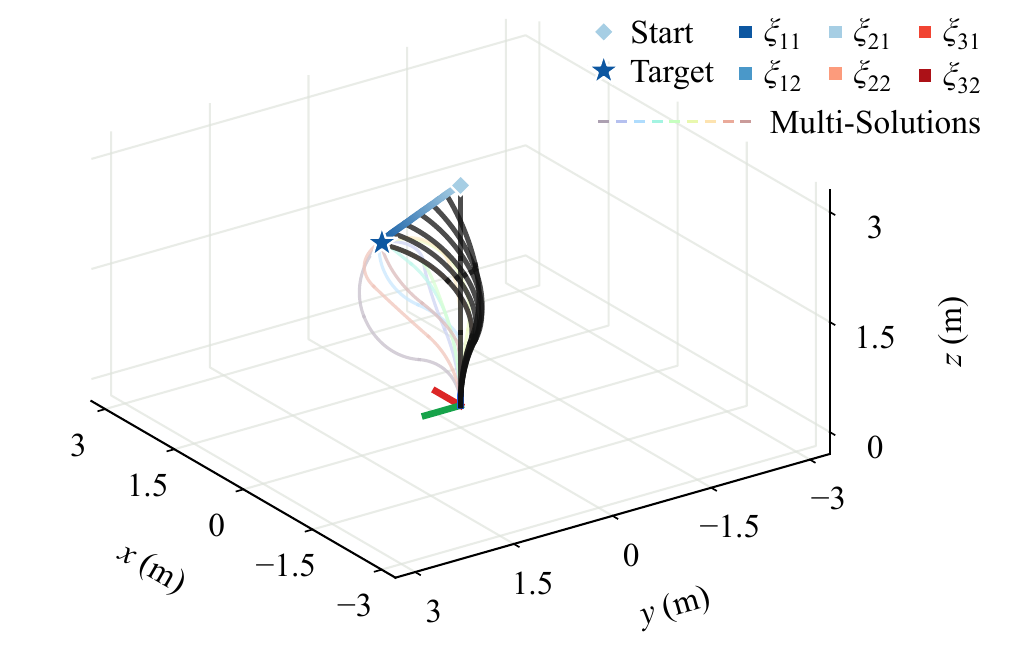}\label{fig:multisolution_a}}\\
  \subfloat[]{\includegraphics[width=1.7in]{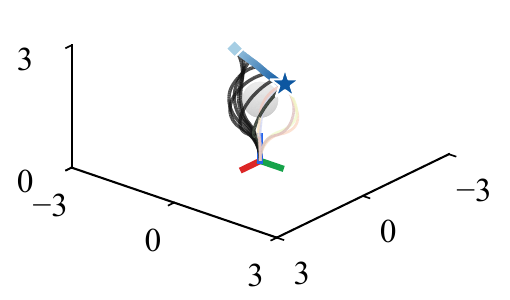}\label{fig:multisolution_b}}
  \subfloat[]{\includegraphics[width=1.7in]{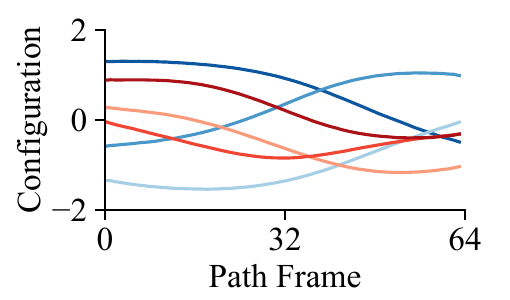}\label{fig:multisolution_c}}\\
  \subfloat[]{\includegraphics[width=1.7in]{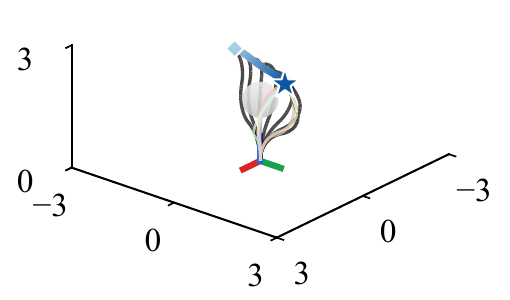}\label{fig:multisolution_d}}
  \subfloat[]{\includegraphics[width=1.7in]{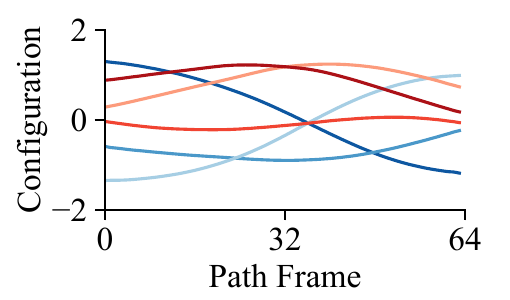}\label{fig:multisolution_e}}\\
  \subfloat[]{\includegraphics[width=1.7in]{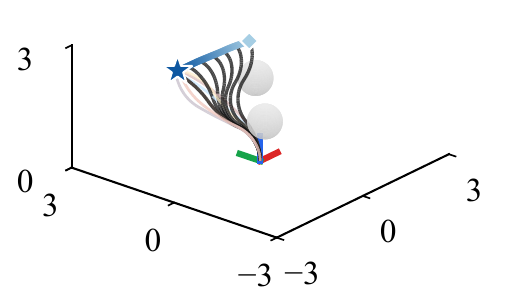}\label{fig:multisolution_f}}
  \subfloat[]{\includegraphics[width=1.7in]{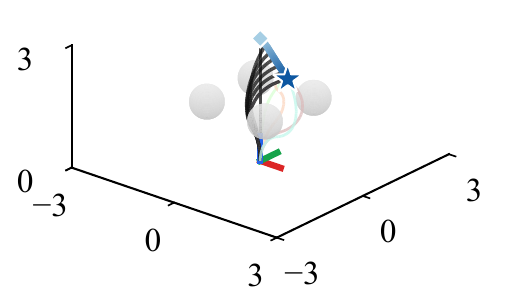}\label{fig:multisolution_g}}\\
  \caption{Multi-solution planning results generated by PccDiffuser in scenes containing zero to four ball obstacles. The examples show alternative terminal configurations, distinct collision-free motions and their corresponding 6-dimensional PCC configuration paths for an identical case, and feasible solutions in increasingly cluttered scenes.}
  \label{fig:multisolution}
\end{figure}

The redundancy of continuum robots permits substantially different configurations and motions to reach the same target. We therefore investigated whether PccDiffuser could capture this multi-modal solution space rather than collapsing to a single path. Using the model trained on the mixed obstacle dataset, we arbitrarily selected some test cases containing several ball obstacles. For each case, pure diffusion without analytical guidance generated 10 configuration-space paths from distinct initial noise samples. We used deterministic DDIM sampling with 50 steps and $\eta=0$, and every path contained a sequence of 6-dimensional PCC configurations. As shown in Fig.~\ref{fig:multisolution}, all samples in the obstacle-free example reached the target within the prescribed tolerance while converging to visibly different terminal configurations. The illustrated cases with different number of obstacles contained multiple feasible samples. Two solutions generated for the same one-obstacle case exhibit different whole-body motions and configuration profiles despite sharing the same initial configuration, target, and environment, while the selected paths in the more cluttered scenes also reach their targets without collision. These examples demonstrate that varying the initial noise enables the learned conditional distribution to represent multiple configuration-space solutions and retain this diversity as the environment becomes more constrained.

\subsection{Ablation Study}

We ablate the analytical guidance discussed in Section~\ref{subsec:sampling_guided_by_differential_kinematics} to quantify its effect on planning accuracy, reliability, and computational cost.

\textbf{Setup.} We compared three variants during the denoising process: the learned diffusion model without analytical guidance, post correction on the predicted terminal configuration, and gradient guidance on predicted noise. All variants used deterministic sampling with 50 steps and $\eta=0$, and were evaluated on the same test dataset with identical initial noise. The planner denoised 10 candidate paths in each case. Analytical refinement was activated during the final 20 denoising steps. We evaluated the variants separately on the obstacle-free dataset and on the obstacle-aware dataset containing zero to four ball obstacles. A path was successful when its terminal tip error was below the tolerance and its sampled backbone remained collision-free throughout the motion. Terminal tip error is reported for successful paths and normalised by the total robot length. Computation was performed on an AMD EPYC 9224 processor with an NVIDIA GeForce RTX 5090.

\begin{figure}[t]
  \centering
  \subfloat[]{\includegraphics[width=3.4in]{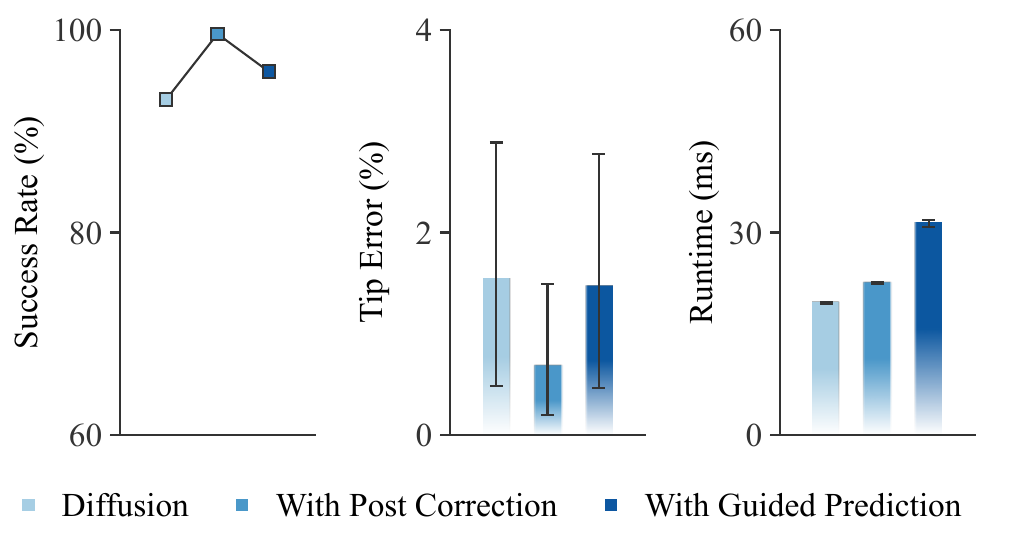}\label{fig:ablation_a}}\\
  \subfloat[]{\includegraphics[width=3.4in]{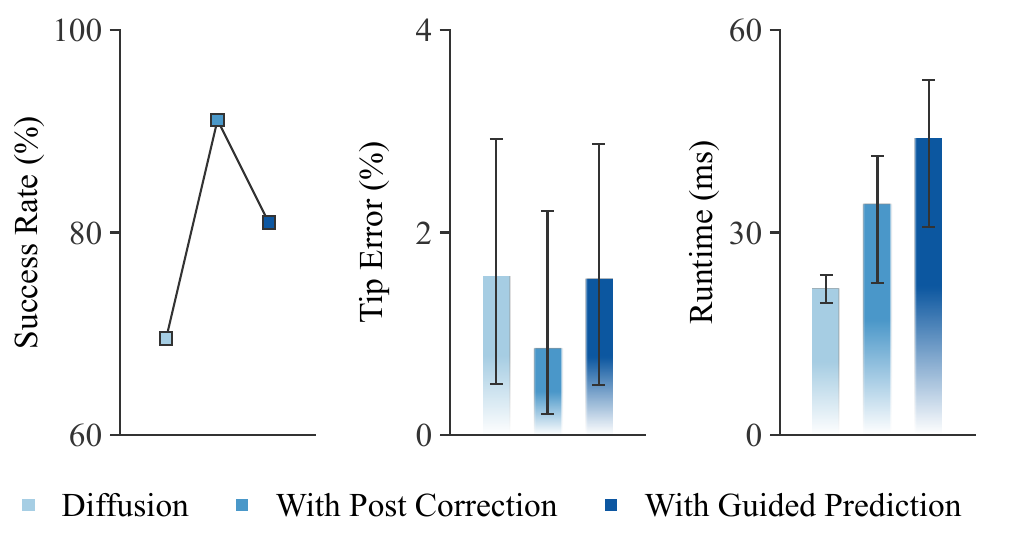}\label{fig:ablation_b}}
  \caption{Ablation of analytical refinement on (a) the obstacle-free dataset and (b) the mixed obstacle dataset containing zero to four ball obstacles. Each panel reports the success rate, terminal tip error normalised by the total robot length, and average runtime per path. Error bars denote the 5th and 95th percentiles.}
  \label{fig:ablation}
\end{figure}

\textbf{Results.} As shown in Fig.~\ref{fig:ablation}, relative to pure diffusion, post correction increased the mean runtime by 15.1\% in an obstacle-free space and 56.8\% in mixed obstacle spaces, whereas guided prediction increased it by 60.4\% and 102.2\%, respectively. Despite this overhead, post correction substantially improved planning performance in both settings. It achieved the highest success rates of 99.62\% and 91.11\%, while reducing the mean normalised terminal tip error by 55.6\% and 45.4\%, respectively. Guided prediction also improved the success rate in both settings, but yielded only marginal gains in tip accuracy. Overall, post correction provided a better trade-off between accuracy and efficiency than guided prediction.

\subsection{Planner Comparison}

We compare our PccDiffuser with representative sampling- and optimisation-based planners to assess whether learning a path distribution improves both planning reliability and computational efficiency.

\textbf{Setup.} We compared PccDiffuser with configuration-space RRT and RRT*, workspace RRT and RRT*, and APF. The configuration-space methods sample PCC configurations directly, whereas the workspace methods sample tip positions and map each extension into configuration space using differential kinematics. The APF benchmark first solves the terminal inverse kinematics, constructs a straight workspace path, and repairs collisions through whole-body repulsion. All methods were tested on the same test dataset containing zero to four ball obstacles. For RRT/RRT*, we used 100 iterations. An attempt was successful only if it returned a path with terminal tip error smaller than the given tolerance, and every sampled backbone point remained collision-free. PccDiffuser processed the test cases one by one, and sampled and denoised 10 candidate solutions in parallel within each case. The benchmark algorithms ran faster on the CPU, so we consequently report the better measured results.

\begin{figure}[t]
  \centering
  \includegraphics[width=3.4in]{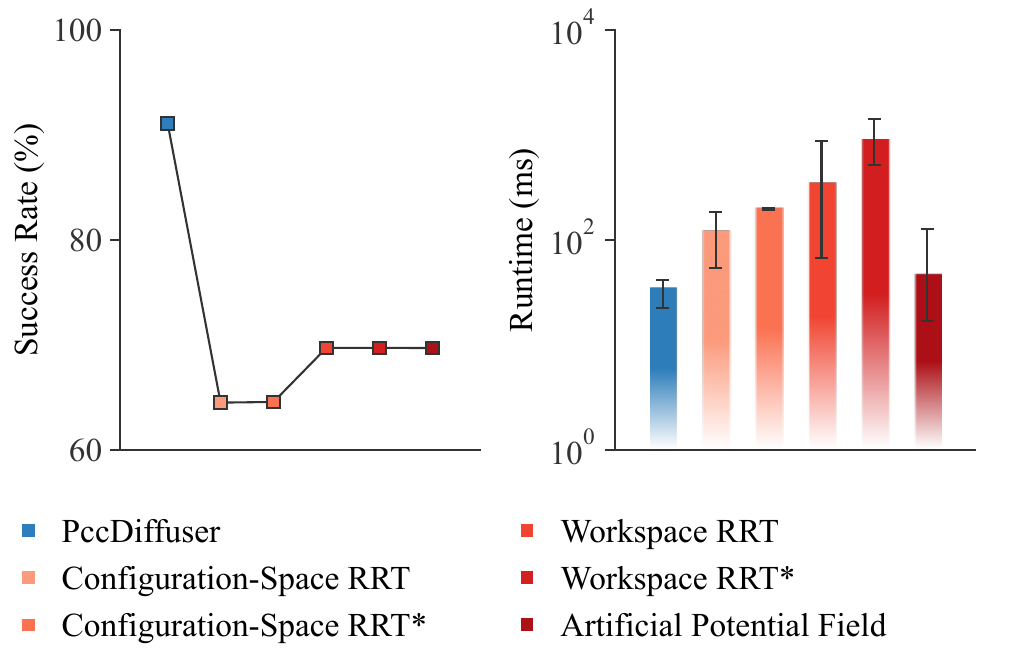}
  \caption{Comparison of success rate (left) and runtime (right) (log scale) between PccDiffuser and benchmark algorithms. Error bars denote the 5th and 95th percentiles.}
  \label{fig:comparison}
\end{figure}


\textbf{Results.} As shown in Fig.~\ref{fig:comparison}, PccDiffuser achieved a success rate of 91.11\%, exceeding the strongest benchmarks, including workspace RRT/RRT* and APF, which achieved a success rate around 70\%. The success rates of configuration-space RRT/RRT* achieved approximately 65\%. PccDiffuser also had the lowest mean runtime of 34.0~ms for each successful path, whereas the runtimes of the other benchmarks were substantially higher. We further found that, because the candidates were denoised in parallel, increasing their number barely changed the total sampling time per planning case, magnifying the advantage of PccDiffuser for multi-solution planning. Within the fixed iteration budget, RRT* increased runtime without materially improving success. These results indicate that the learned path prior guides samples towards feasible regions more reliably than uninformed exploration, and that batched denoising provides a lower average runtime than the iterative benchmarks.

\subsection{Real-World Evaluation}

We deployed the proposed planner on a three-section tendon-driven continuum robot to examine whether the planning results remain executable on our hardware platform. In each case we generated multiple trajectories in the configuration space, from which one was selected based on test-time requirements and interpolated in the controller for execution. We evaluated three capabilities: consecutive planning to targets received online, multiple solutions to an identical target tip position, and whole-body obstacle avoidance.

\begin{figure}[t]
  \centering\vspace{-10pt}
  \subfloat[]{\includegraphics[width=3.4in]{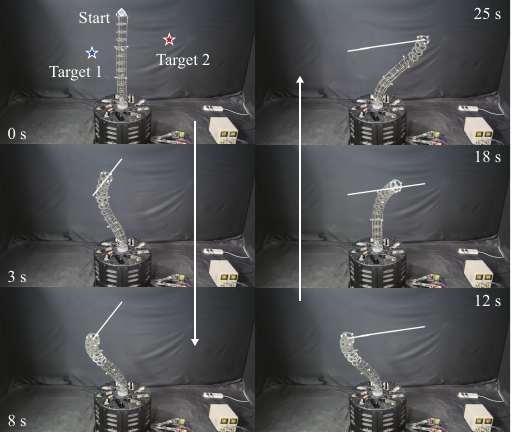}\label{fig:real_world_consecutive_a}}\\
  \subfloat[]{\includegraphics[width=3.4in]{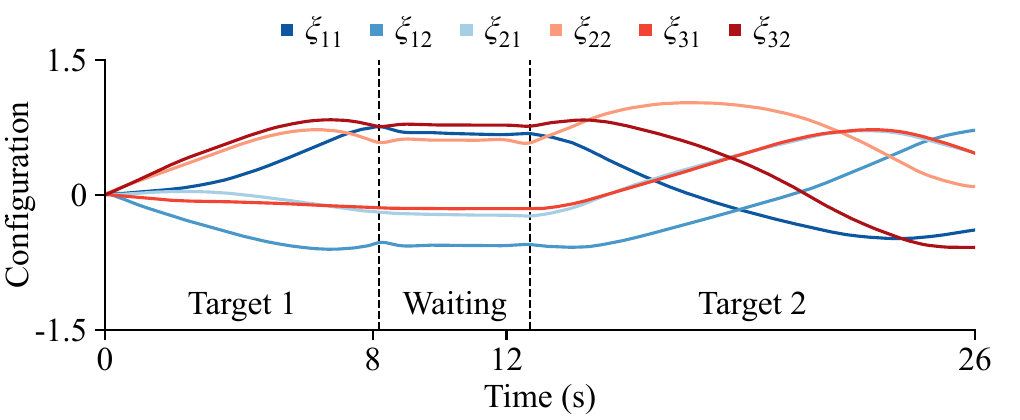}\label{fig:real_world_consecutive_b}}
  \caption{Consecutive planning to two target points. (a) Hardware snapshots as the robot moves from the zero configuration to Target~1, holds the state for a while, and then moves to Target~2. (b) Measured configurations during execution. The black dashed lines separate the first motion, waiting, and second motion stages. (See also the supplementary video.)}
  \label{fig:real_world_consecutive}
\end{figure}

\textbf{Consecutive Planning.} The robot started from the zero configuration and was first commanded to the normalised workspace target $(0,-1.55,2.25)$. We generated 10 candidates and executed one sample. The terminal configuration then served as the initial condition for a second case to the target $(1.5,0,2.4)$. As shown in Fig.~\ref{fig:real_world_consecutive_a}, the robot reached the first target, maintained its shape during the waiting interval, and subsequently moved to the second target. The measured configurations in Fig.~\ref{fig:real_world_consecutive_b} remain continuous across both replanning boundaries, confirming that conditioning the second target on the intermediate state enables smooth and consecutive execution.

\begin{figure}[t]
  \centering\vspace{-10pt}
  \subfloat[]{\includegraphics[width=3.4in]{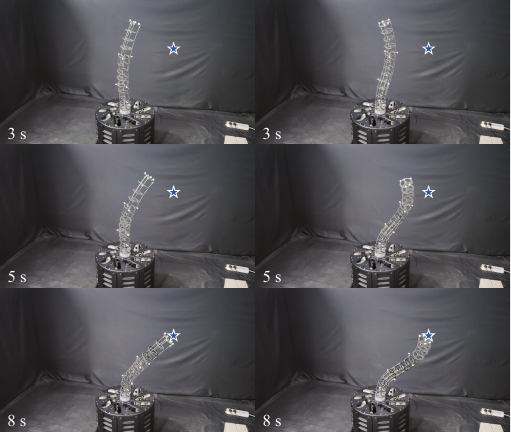}\label{fig:real_world_multisolution_a}}\\
  \subfloat[]{\includegraphics[width=3.4in]{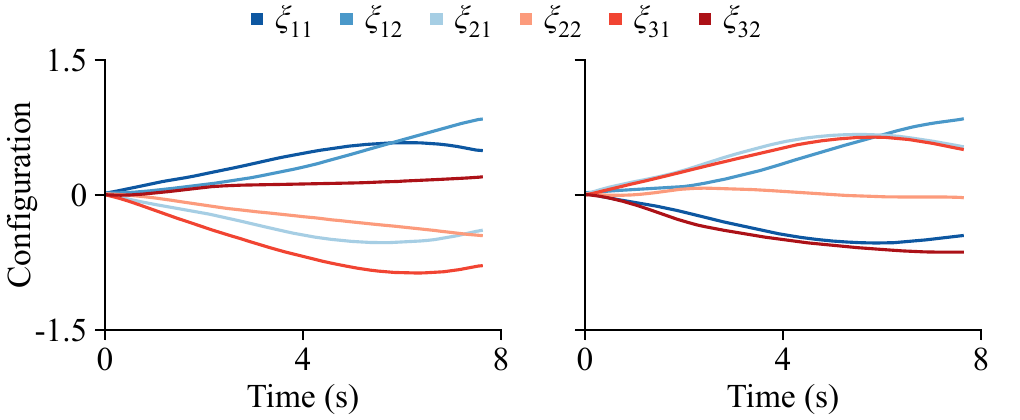}\label{fig:real_world_multisolution_b}}
  \caption{Two distinct planning solutions for identical start configuration and target position. (a) Hardware snapshots for the two samples (left and right). Both of them reach the target (star) through different backbone shapes. (b) Measured configurations for each solution, showing two distinct configuration-space trajectories. (See also the supplementary video.)}
  \label{fig:real_world_multisolution}
\end{figure}

\textbf{Multi-Solution Planning.} To test the multi-solution capability of our planner, we fixed the zero initial configuration and the normalised workspace target tip position $(1.55,-0.07,2.33)$, and executed two samples from the 10 candidates in separate trials. Both motions reached the marked target with visibly different whole-body shapes, as shown in Fig.~\ref{fig:real_world_multisolution_a}. Their measured configuration histories in Fig.~\ref{fig:real_world_multisolution_b} are likewise well separated. These results confirm that the sampled solutions are not online perturbations of one nominal path. They provide alternative motions for the same boundary conditions.

\textbf{Obstacle Avoidance Planning.} Finally, we placed a ball obstacle of normalised radius $0.4$ at $(0,0,1.5)$. The planner first generated motion from the zero configuration to target tip position $(-1.45,0,2.25)$. Then, starting from its terminal configuration, a batch of trajectories was generated to $(1.35,0,2.45)$ and one of the solutions was selected. Collision checking considered the complete discretised backbone rather than only the tip. Fig.~\ref{fig:teaser} shows how the initially dispersed path points evolve during denoising and converge to the obstacle-aware solution. The executed path retained whole-body clearance from the obstacle and reached the target without collisions, demonstrating that the learned obstacle condition and geometric correction remain effective on the hardware platform.

\section{CONCLUSION}

In this paper, we presented the PccDiffuser, a conditional diffusion framework for multi-solution motion planning of continuum robots. The kinematic parameterisation supports path generation around the straight configuration, while analytical guidance improves terminal accuracy and whole-body clearance during the denoising process. Time allocation under actuator constraints turns selected paths into executable trajectories. Simulation results showed that the proposed planner generates diverse solutions in both free and cluttered spaces, with post correction offered the best balance between planning success, tip accuracy, and computational cost. Our planner outperformed conventional sample- and optimisation-based benchmarks. Hardware experiments further validated the proposed planner. We release our code at \url{https://github.com/qiuke-qiuke/pcc_diffuser}. Future work will investigate dynamic environments and closed-loop replanning with online feedback.





\bibliographystyle{IEEEtran}
\bibliography{ref}


\end{document}